\documentclass{article}
\usepackage{amssymb}
\usepackage{amsmath}
\usepackage{graphicx}
\usepackage{float} 
\usepackage{multirow} 
\usepackage[final]{corl_2025} % Uncomment for the camera-ready ``final'' version.

\title{Improving the performance of AI-powered Affordable Robotics for Assistive Tasks}

\author{
Dharunish Yugeswardeenoo\\
\textit{Central Bucks High School South, Warrington, PA, United States}\\
\texttt{dharyugi@gmail.com}
}
\begin{document}
\maketitle

%===============================================================================

\begin{abstract}
By 2050, the global demand for assistive care is expected to reach 3.5 billion people, far outpacing the availability of human caregivers. Existing robotic solutions remain expensive and require technical expertise, limiting accessibility. This work introduces a low-cost robotic arm for assistive tasks such as feeding, cleaning spills, and fetching medicine. The system uses imitation learning from demonstration videos, requiring no task-specific programming or manual labeling. The robot consists of six servo motors, dual cameras, and 3D-printed grippers. Data collection via teleoperation with a leader arm yielded 50,000 video frames across the three tasks. A novel Phased Action Chunking Transformer (PACT) captures temporal dependencies and segments motion dynamics, while a Temporal Ensemble (TE) method refines trajectories to improve accuracy and smoothness. Evaluated across five model sizes and four architectures, with ten hours of real-world testing, the system achieved over 90\% task accuracy, up to 40\% higher than baselines. PACT enabled a 5× model size reduction while maintaining 75\% accuracy. Saliency analysis showed reliance on key visual cues, and phase token gradients peaked at critical trajectory moments, indicating effective temporal reasoning. Future work will explore bimanual manipulation and mobility for expanded assistive capabilities.
\end{abstract}

\keywords{Imitation learning, Assistive tasks, Open Source Robot Arm} 

%===============================================================================

\section{Introduction}
	
An estimated 1.3 billion people worldwide live with significant disabilities \citep{who2023disability}. In the United States, 46\% of people 75 and older report a disability \citep{uscensusdata}, and 23.3 million are unable to live independently \citep{cdc2024infographic}, highlighting a substantial need for accessible assistive technologies. By 2050, the global population aged 60 and older is projected to double to 2.1 billion \citep{who2023disability} further increasing the demand for human-assisted care and exacerbating the caregiver-to-recipient imbalance. Robotic assisted living offers a scalable solution to address caregiver shortages and support the aging population. However, existing robotic systems are often prohibitively expensive, ranging from \$20,000 to \$400,000 \citep{ackerman2023stretch}, and require specialized expertise for construction and programming.
This research aims to create a low-cost, AI-driven robotic arm capable of assisting disabled individuals to complete everyday tasks independently. The robot leverages an approach of imitation learning \citep{gavenski2024survey}, in which robots learn from observing and mimicking human actions without needing complex programming or in-depth knowledge of the robot mechanics. 

\section{Methodology}

\textbf{Robot Arm and Testing Environment.} This work uses the LeRobot framework \citep{cadene2024lerobot}. LeRobot is an open source framework that provides build instructions for the robot arm, baseline AI models, configuration files, and teleoperation scripts. The primary robot arm is a variant of the Moss robot arm \citep{mossrobotarms} that features six motors, offering six degrees of freedom, and is constructed from brackets, screws, motors, and a 3D-printed gripper. A second, structurally identical arm is used as a secondary arm to enable teleoperation. The secondary arm has a 3D printed handle instead of gripper. Experimental data is recorded using a smartphone mounted on a tripod for top-view and a laptop for side-view capture. The AI model runs on the laptop and communicates with the arm via USB connection to the motor control board. Data collection is performed via teleoperation, where the primary (follower) arm replicates the movements of a secondary (leader) arm manually controlled by a human. Dataset captures the synchronized top-view and side-view videos of the robotic arm performing tasks, along with recorded motor angles at each timestep.

\begin{figure}[H]
     \centering
     \includegraphics[width=1.0\linewidth]{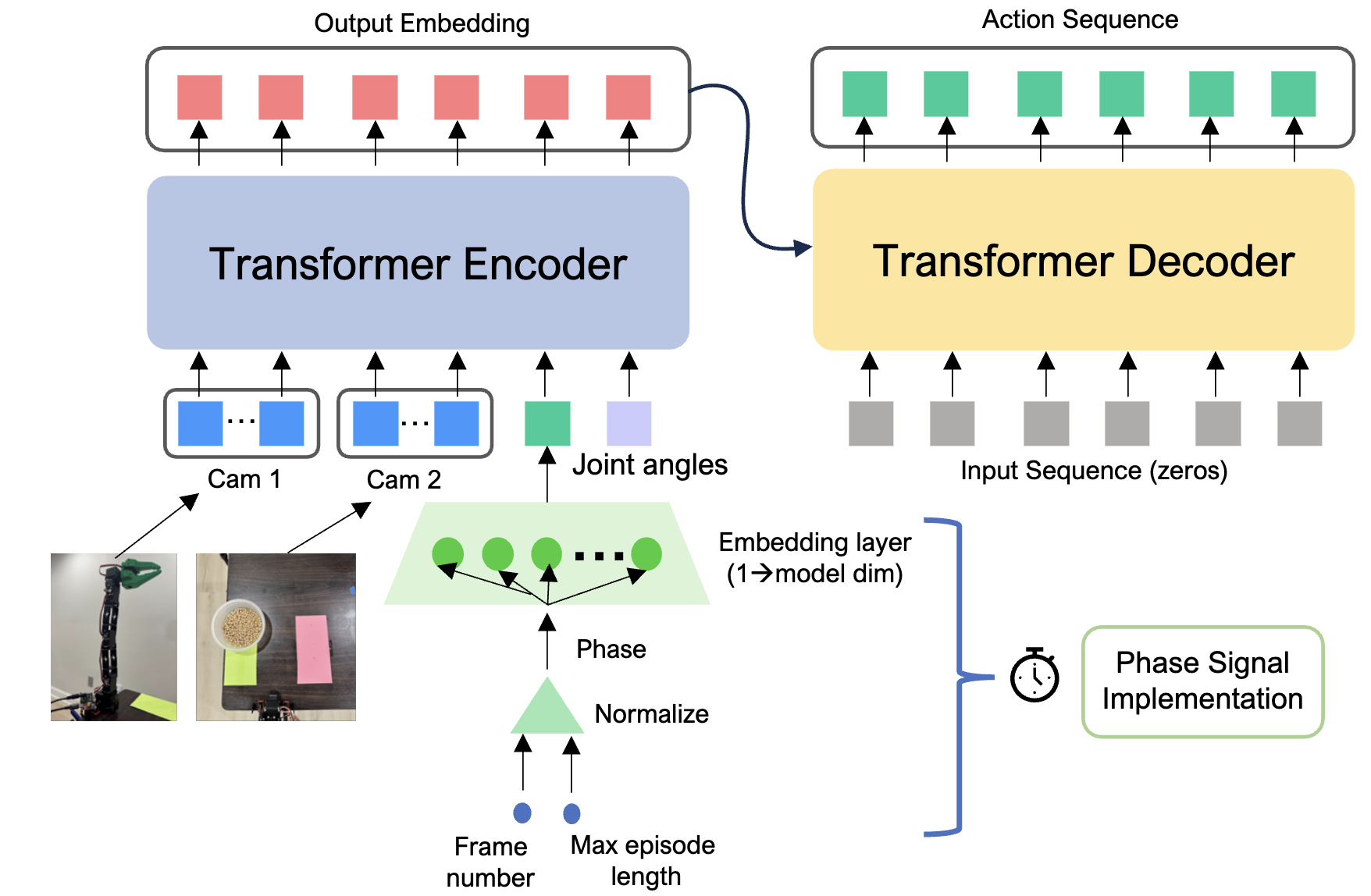}
     \caption{PACT Architecture}
     \label{fig:PACT}
 \end{figure}

\textbf{Phased Action Chunking Transformer Architecture (PACT)}. This work adopts the state-of-the-art Action Chunking Transformer (ACT) \citep{zhao2023bimanual} architecture as the experimental baseline. The ACT model aims to solve challenges of imitation learning, such as compounding errors and non-Markovian behaviors, by predicting actions in "chunks". ACT when applied to assistive tasks reveals a lack of temporal awareness, leading to poor performance in visually ambiguous scenarios. Specifically, the model tends to produce similar actions for visually similar frames, even when the appropriate actions differ. To address this limitation, this work proposes a novel Phased Action Chunking Transformer (PACT) (Figure \ref{fig:PACT}), an extension of ACT that incorporates phase information into the architecture to better model temporal context. We introduce a phase token that acts as a learnable temporal clock, consisting of a normalization between the current frame number and the maximum episode length. This encourages the model to disambiguate similar-looking frames by when they occur. We utilize the same encoder/decoder as ACT.

\begin{figure}[H]
    \centering
    \includegraphics[width=1.0\linewidth]{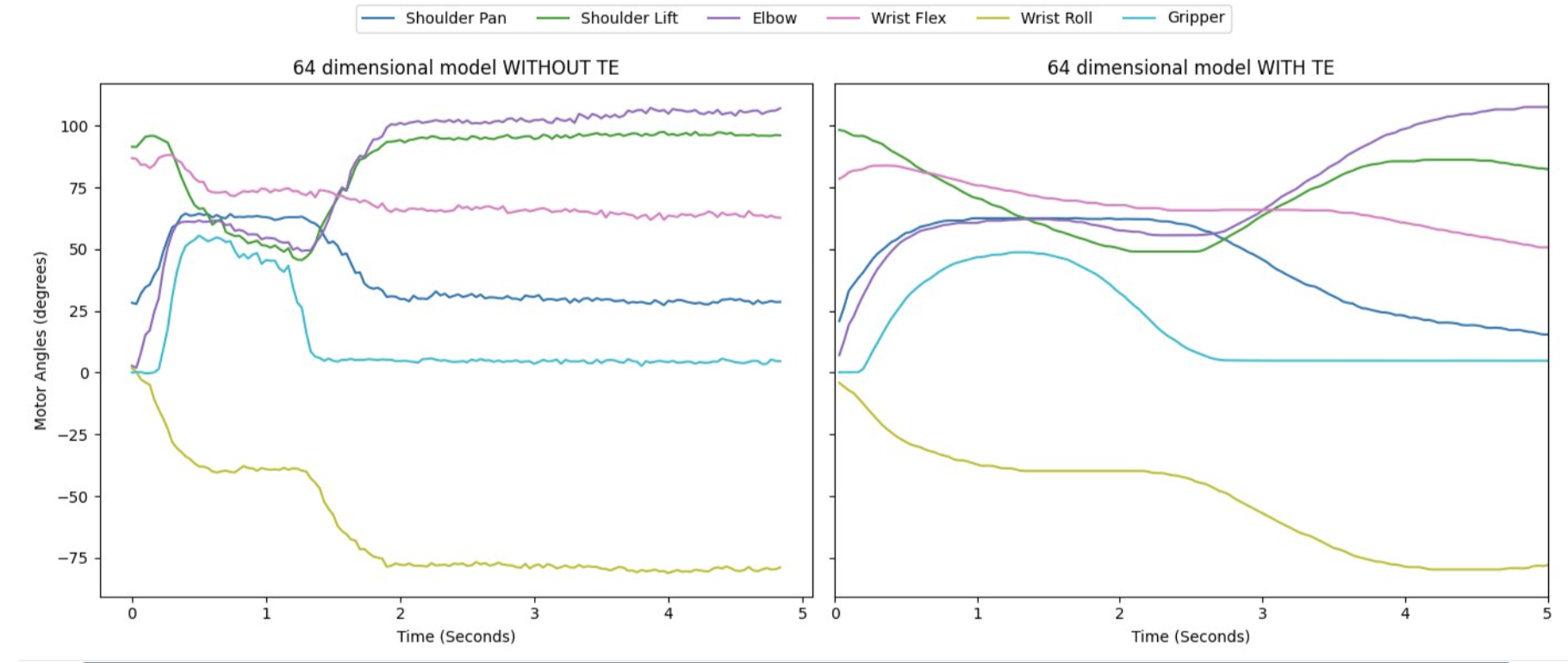}
    \caption{Visual of TE's impact is shown in terms of the evolution of motor angles over a time. The 64-dimensional model without TE (left) outputs jittery and inconsistent movement. The right graph shows the same model but with TE applied, displaying smoother and slower motion. Both models are examined on the same test case.}
    \label{fig:TEvis}
\end{figure}

\textbf{Temporal Ensemble (TE)}. The impact of embedding dimension on the performance is evaluated. Baseline ACT models with embedding dimensions of 64 or lower leads to jittery, inconsistent motion. To overcome this limitation, we employ temporal ensembling \citep{zhao2023bimanual}, where the model generates predictions at every timestep and overlapping outputs are averaged. This approach yields smoother and more stable trajectories. While it increases computational load and slows robot movements, the trade-off is acceptable, particularly in real-world assistive settings, where safety and reliability around humans take priority over speed. The effect of TE can be shown in Figure \ref{fig:TEvis}.

\section{Experimental Setup}
\begin{figure}
    \centering
    \includegraphics[width=0.95\linewidth]{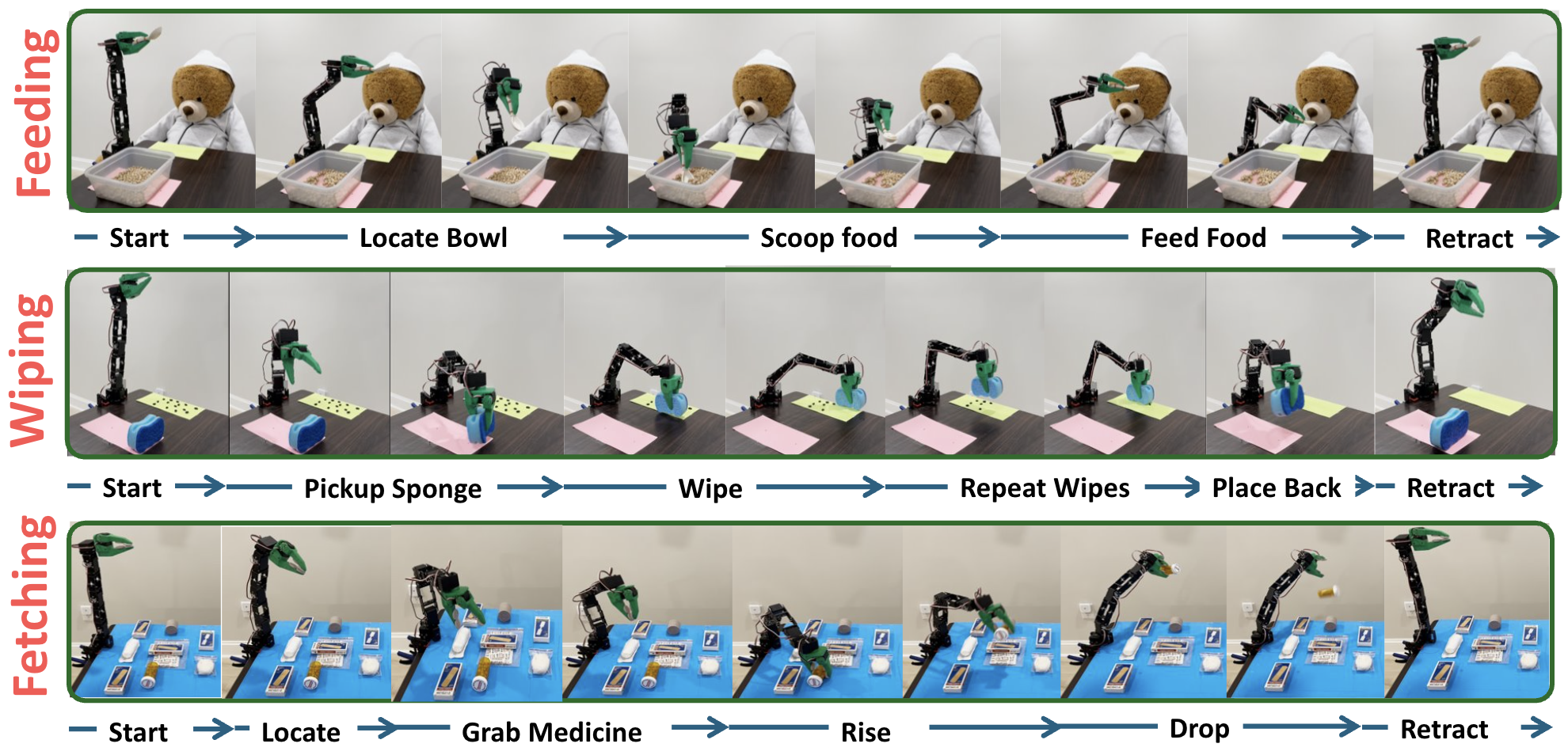}
    \caption{Three different assistive tasks and respective sub-tasks.}
    \label{fig:taskVis}
\end{figure}

\textbf{Task Definition. }The tasks evaluated are shown in Figure \ref{fig:taskVis}.

\textbf{Data Collection.} In total, via teloperation, 39 demonstrations were collected (18,000 frames), the wiping task involved 47 demonstrations (27,000 frames), and the fetching medicine task used 45 demonstrations (13,500 frames). During dataset creation, different configurations of objects (e.g. color, placement,) were placed to ensure generalization.

\textbf{Evaluation.} A total of 30 models were used during experimentation, including 5 different model embedding dimensions: 32, 64, 128, 256, and 512; 4 different architectures:  ACT, PACT, ACT+TE, PACT+TE; 3 different tasks: Feeding, Wiping, and Fetching Medicine. For each combination, accuracy of task completion is measured.

\section{Results}
Results for the feeding, wiping, and fetching medicine tasks are shown in Table \ref{tab:allAcc}. PACT+TE architecture improves performance of low embedding size models consistently across all tasks.

\begin{table}[h]
    \centering
    \resizebox{\textwidth}{!}{%
    \begin{tabular}{|c|c|c|c|c||c|c|c|c||c|c|c|c|}\hline
        \multirow{2}{*}{Dim} 
        & \multicolumn{4}{c||}{Feeding} 
        & \multicolumn{4}{c||}{Wiping} 
        & \multicolumn{4}{c|}{Fetching} \\\cline{2-13}
        & ACT & ACT+TE & PACT & PACT+TE 
        & ACT & ACT+TE & PACT & PACT+TE 
        & ACT & ACT+TE & PACT & PACT+TE \\\hline
        32  & \_   & 67.2 & \_   & \textbf{76.1}  & \_   & 59.6 & \_   & \textbf{75.9}  & \_   & 9.9  & \_   & \textbf{63.2} \\\hline
        64  & \_   & 72.1 & \_   & \textbf{85.2}  & \_   & 88.1 & \_   & \textbf{96.3}  & \_   & 16.6 & \_   & \textbf{70.0} \\\hline
        128 & 54.8& 73.1 & 73.1& \textbf{90.8}  & 97.5& 91.2 & 98.3& \textbf{100.0} & 50.0& 63.2 & 73.3& \textbf{73.3} \\\hline
        256 & 66.4& \textbf{87.6} & 69.1& 87.1  & 97.6& \textbf{99.8} & 98.9& 99.6  & 69.9& 76.6 & \textbf{83.3}& 80.0 \\\hline
        512 & 68.4& 76.5 & 69.7& \textbf{88.1}  & 99.0& 100.0& 99.8& \textbf{100.0} & 79.9& 86.6 & 93.3& \textbf{100.0} \\\hline
    \end{tabular}
}

    \caption{Accuracy results for Feeding, Wiping, and Fetching Medicine tasks across different model sizes. Note: Low dimensional models (32,64) are unstable without TE.}
    \label{tab:allAcc}
\end{table}

\section{Saliency Analysis}

\begin{figure}[H]
    \centering
    \includegraphics[width=1.0\linewidth]{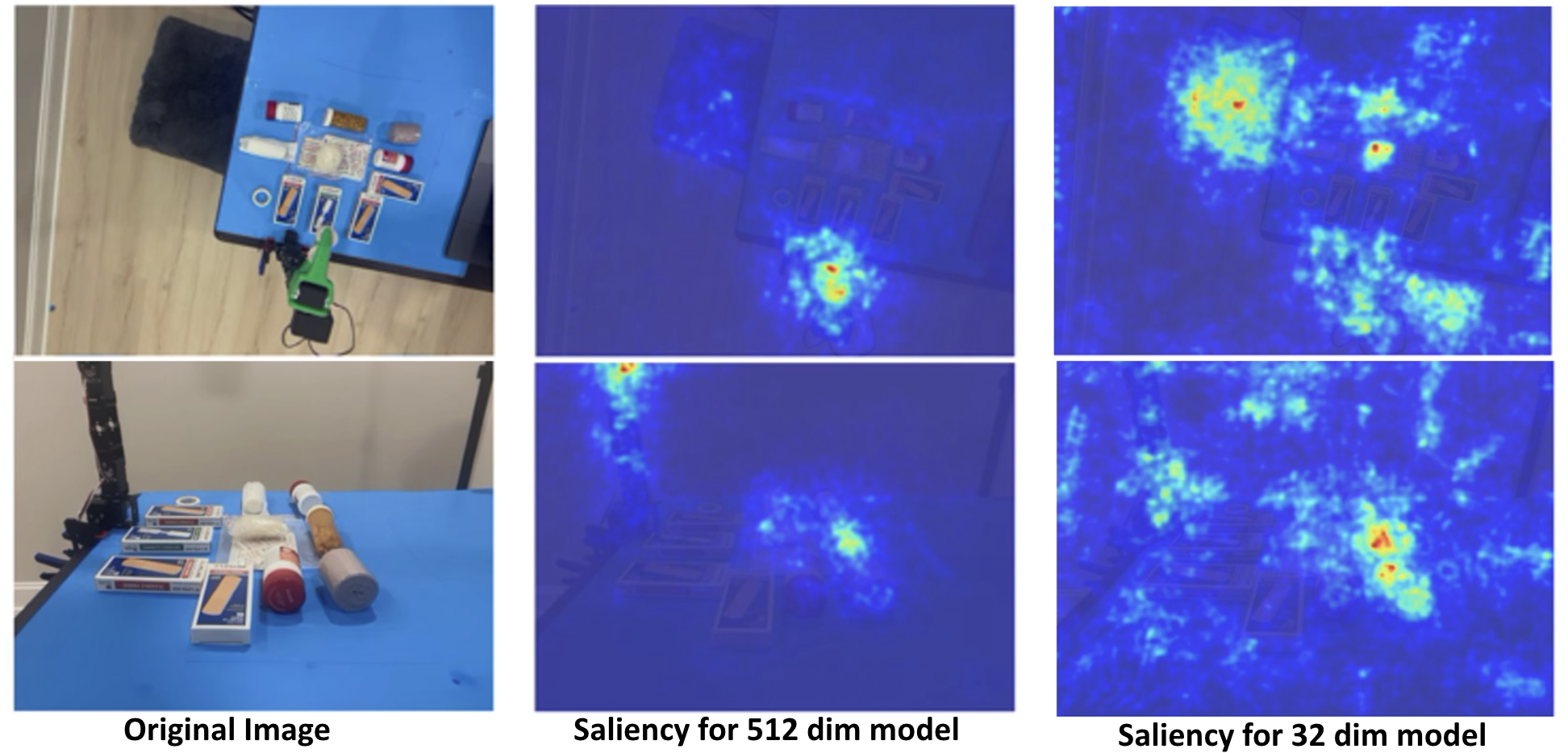}
    \caption{The above figure highlights the saliency for the input images for the 512 dimensions and 32 dimensional models on a test case in the fetching medicine task.}
    \label{fig:imgSal}
\end{figure}

\textbf{Image Token Analysis.} Figure \ref{fig:imgSal} illustrates saliency maps for the 512-dimensional and 32-dimensional models during a test case in the fetching medicine task. The 512-dimensional model exhibits concentrated attention on both the medicine container and the robot arm, as indicated by prominent warm regions. In contrast, the 32-dimensional model displays more diffuse saliency, spreading attention across a broader range of pixels, including irrelevant areas. This supports the assertion that lower-dimensional representations lack the precision necessary for accurate predictions. Their broader focus introduces noise, impairing the model's ability to grasp and transport the container stably. Interestingly, the model tends to ignore static background elements that remain unchanged during training. 

\begin{figure}[H]
    \centering
    \includegraphics[width=1.0\linewidth]{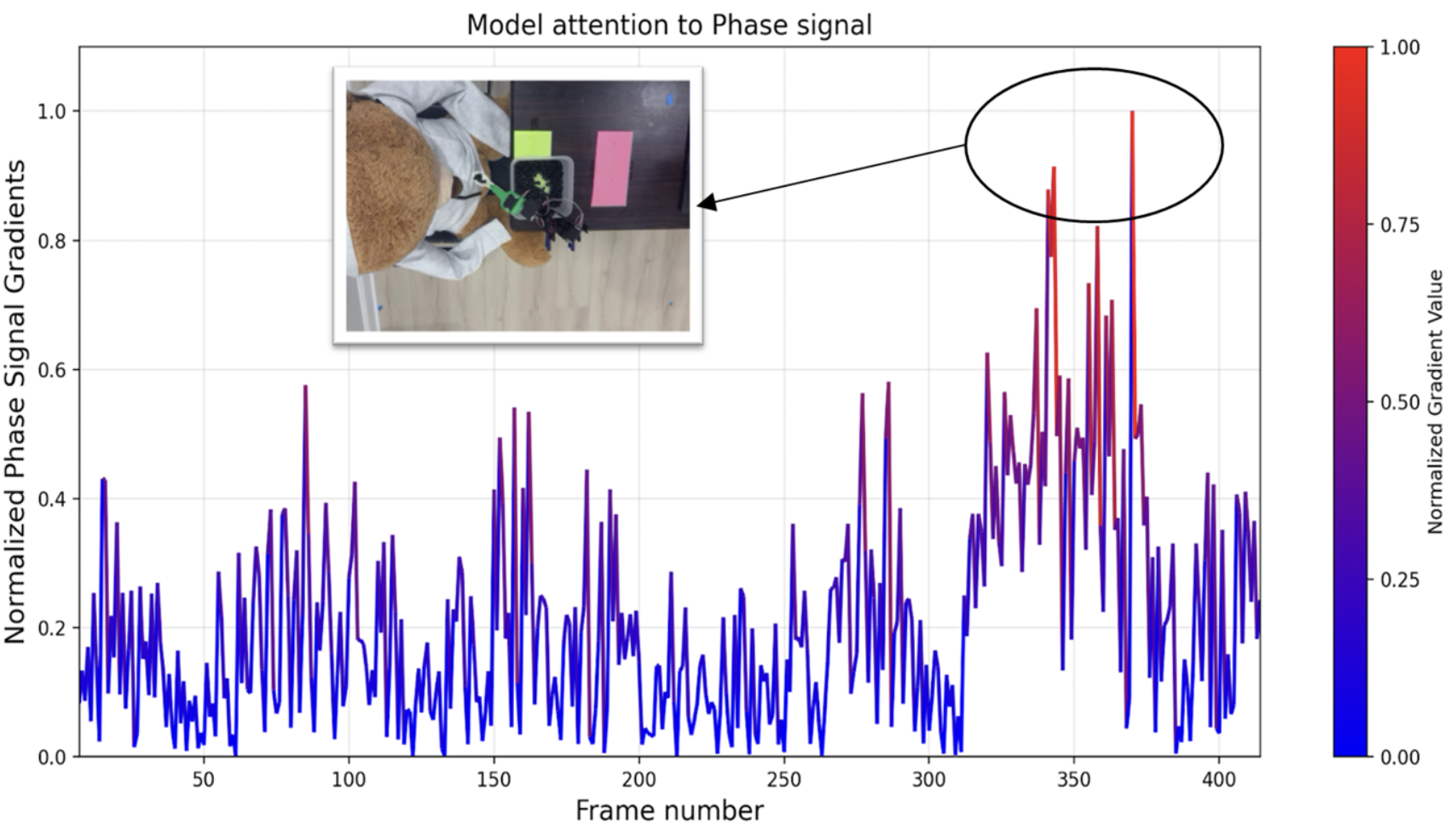}
    \caption{The above figure displays the saliency of the phase token at each frame in an episode of the fetching medicine task. The model's attention towards the phase token peaks at an instance where visual ambiguity was frequently observed.}
    \label{fig:PACTsal}
\end{figure}

\textbf{Phase Token Analysis.} Saliency analysis was also applied to the PACT phase token by computing the gradient of the phase signal at each frame. Figure \ref{fig:PACTsal} illustrates these gradients over time for a representative example from the feeding task. The model demonstrates dynamic attention to the phase token across the episode, with a significant peak around frame 350 corresponding to the moment the robot extended its arm and prepared to retract. Although the robot successfully completed the task, this point coincided with challenges related to repetition and confusion caused by visually similar frames. 

\section{CONCLUSION}

AI can significantly enhance the performance of low-cost robots, making them more affordable and viable for widespread use in assisted living facilities. Imitation learning enables easy robot skill acquisition without programming, task-specific pipelines, or manual labeling. The PACT (Phased-Action Chunking Transformer) architecture is a promising solution for accuracy and efficiency improvement. PACT allows reducing model size without significantly impacting performance. PACT reduces computational requirements and overall cost, making AI-enabled robotics more feasible for cost-conscious applications.In addition, Temporal Ensembling plays a vital role in motion trajectory smoothing for low-dimensional models. By removing jitter and inconsistencies, this approach ensures smooth, stable, and safe robotic motions crucial for assistive tasks.

\clearpage
% The acknowledgments are automatically included only in the final and preprint versions of the paper.
%\acknowledgments{If a paper is accepted, the final camera-ready version will (and probably should) include acknowledgments. All acknowledgments go at the end of the paper, including thanks to reviewers who gave useful comments, to colleagues who contributed to the ideas, and to funding agencies and corporate sponsors that provided financial support.}

%===============================================================================

\bibliography{example}  % .bib

\section{Appendix}

\subsection{PACT Architecture}
\subsection{Temporal Ensembling}
\subsection{Image Saliency}
\subsection{PACT Saliency}
\end{document}